\documentclass[10pt,twocolumn,letterpaper]{article}

\usepackage{cvpr}              

\usepackage{graphicx}
\usepackage{amsmath}
\usepackage{amssymb}
\usepackage{amsthm}
\usepackage{amsfonts}
\usepackage{dsfont}
\usepackage{booktabs}

\usepackage{times}
\usepackage{amsfonts}
\usepackage{bm}
\usepackage{tabu}
\usepackage{multirow}
\usepackage{color}
\usepackage[abs]{overpic}
\usepackage{comment}
\usepackage{textcomp}
\usepackage[accsupp]{axessibility}

\def\etal{\emph{et al.}}
\def\ie{\emph{i.e.},}

\definecolor{mypink}{RGB}{236, 2, 141}

\usepackage[pagebackref,breaklinks,colorlinks]{hyperref}

\usepackage[capitalize]{cleveref}
\crefname{section}{Sec.}{Secs.}
\Crefname{section}{Section}{Sections}
\Crefname{table}{Table}{Tables}
\crefname{table}{Tab.}{Tabs.}

\newcommand*{\affaddr}[1]{#1} 
\newcommand*{\affmark}[1][*]{\textsuperscript{#1}}
\newcommand*{\email}[1]{\small{\texttt{#1}}}

\begin{document}

\title{Joint Visual Grounding and Tracking with Natural Language Specification}

\author{Li Zhou\affmark[1], Zikun Zhou\affmark[2,1*], Kaige Mao\affmark[1], and Zhenyu He\affmark[1,*]\\\affaddr{\affmark[1]Harbin Institute of Technology, Shenzhen}\quad\affaddr{\affmark[2]Peng Cheng Laboratory}\\
\email{lizhou.hit@gmail.com\quad zhouzikunhit@gmail.com\quad maokaige.hit@gmail.com\quad zhenyuhe@hit.edu.cn}\\
}

\maketitle
\renewcommand{\thefootnote}{\fnsymbol{footnote}} 
\footnotetext{$^{*}$Corresponding authors: Zikun Zhou and Zhenyu He.}

\begin{abstract}
Tracking by natural language specification aims to locate the referred target in a sequence based on the natural language description. Existing algorithms solve this issue in two steps, visual grounding and tracking, and accordingly deploy the separated grounding model and tracking model to implement these two steps, respectively. Such a separated framework overlooks the link between visual grounding and tracking, which is that the natural language descriptions provide global semantic cues for localizing the target for both two steps. Besides, the separated framework can hardly be trained end-to-end. To handle these issues, we propose a joint visual grounding and tracking framework, which reformulates grounding and tracking as a unified task: localizing the referred target based on the given visual-language references. Specifically, we propose a multi-source relation modeling module to effectively build the relation between the visual-language references and the test image. In addition, we design a temporal modeling module to provide a temporal clue with the guidance of the global semantic information for our model, which effectively improves the adaptability to the appearance variations of the target. 
Extensive experimental results on TNL2K, LaSOT, OTB99, and RefCOCOg demonstrate that our method performs favorably against state-of-the-art algorithms for both tracking and grounding. Code is available at \href{https://github.com/lizhou-cs/JointNLT}{https://github.com/lizhou-cs/JointNLT}.
\end{abstract}

\section{Introduction}
\begin{figure}[t]
\centering
\includegraphics[width=0.99\columnwidth]{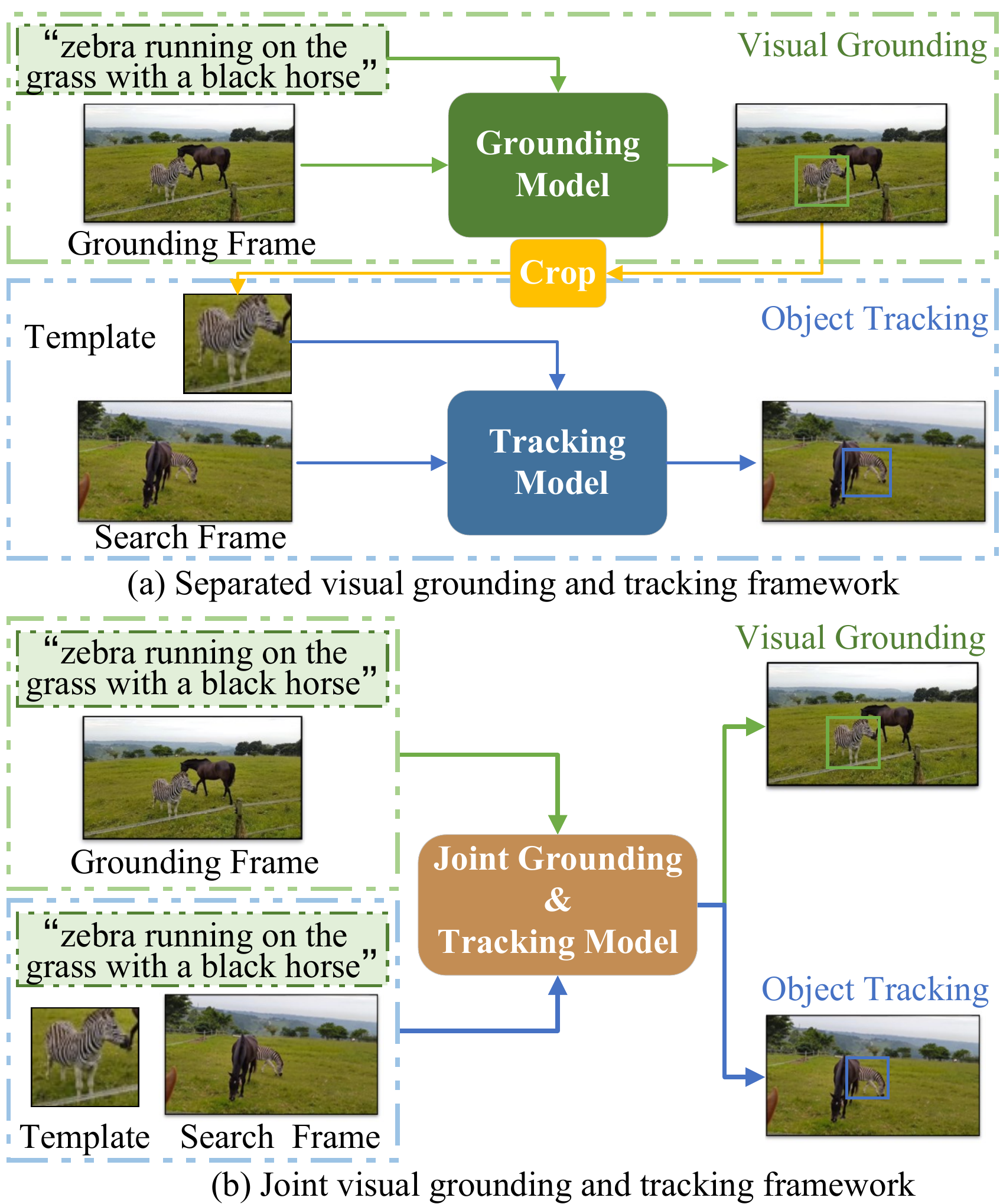}
\vspace{-2mm}
\caption{Illustration of two different frameworks for tracking by natural language specification. (a) The separated visual grounding and tracking framework, which consists of two independent models for visual grounding and tracking, respectively. (b) The proposed joint visual grounding and tracking framework, which employs a single model for both visual grounding and tracking.}
\label{Fig:introduction}
\vspace{-2mm}
\end{figure}
Tracking by natural language specification~\cite{li2017tracking} is a task aiming to locate the target in every frame of a sequence according to the state specified by the natural language.
Compared with the classical tracking task~\cite{yilmaz2006object, OTB2013, OTB2015, TrackingNet} using a bounding box to specify the target of interest, tracking by natural language specification provides a novel human-machine interaction manner for visual tracking. 
In addition, the natural language specification also has two advantages for the tracking task compared to the bounding box specification. 
First, the bounding box only provides a static representation of the target state, while the natural language can describe the variation of the target for the long term. 
Second, the bounding box contains no direct semantics about the target and even results in ambiguity~\cite{TNL2K}, but the natural language can provide clear semantics of the target used for assisting the tracker to recognize the target.
In spite of the above merits, tracking by natural language specification has not been fully explored.

Most existing solutions~\cite{li2017tracking, GTI, TNL2K, li2022cross} for this task could be generally divided into two steps: (1) localizing the target of interest according to the natural language description in the first frame, \ie~visual grounding; (2) tracking the localized target in the subsequent frames based on the target state predicted in the first frame, \ie~visual tracking. 
Accordingly, many algorithms~\cite{GTI, TNL2K, li2022cross} are designed to incorporate a grounding model and a tracking model, as shown in Figure~\ref{Fig:introduction}(a). 
Herein the grounding model performs relation modeling between the language and vision signal to localize the target, while the tracking model performs relation modeling between the template and search region to localize the target. 
The drawback of this framework is that the grounding model and the tracking model are two separate parts and work independently, ignoring the connections between the two steps. 
Besides, many of them~\cite{GTI, TNL2K, li2022cross} choose to adopt the off-the-shelf grounding model~\cite{2019onestagevg} or tracking model~\cite{li2019siamrpn++} to construct their framework, which means that the overall framework cannot be trained end-to-end.

The tracking model in most existing algorithms~\cite{GTI, TNL2K,li2022cross} predicts the target state only based on the template, overlooking the natural language description. 
By contrast, the tracking mechanism that considers both the target template and the natural language for predicting the target state has proven to have great potential~\cite{li2017tracking,wang2018describe, feng2021siamese, guo2022divert}. 
Such a tracking mechanism requires the tracking model to own the ability to simultaneously model the vision-language relation and the template-search region relation. 
Inspired by this tracking mechanism, we come up with the idea to build a joint relation modeling model to accomplish the above-mentioned two-step pipeline. 
Herein a joint relation modeling model can naturally connect visual grounding and tracking together and also can be trained end-to-end.
\begin{figure*}[h]
\centering
\includegraphics[width=0.98\textwidth]{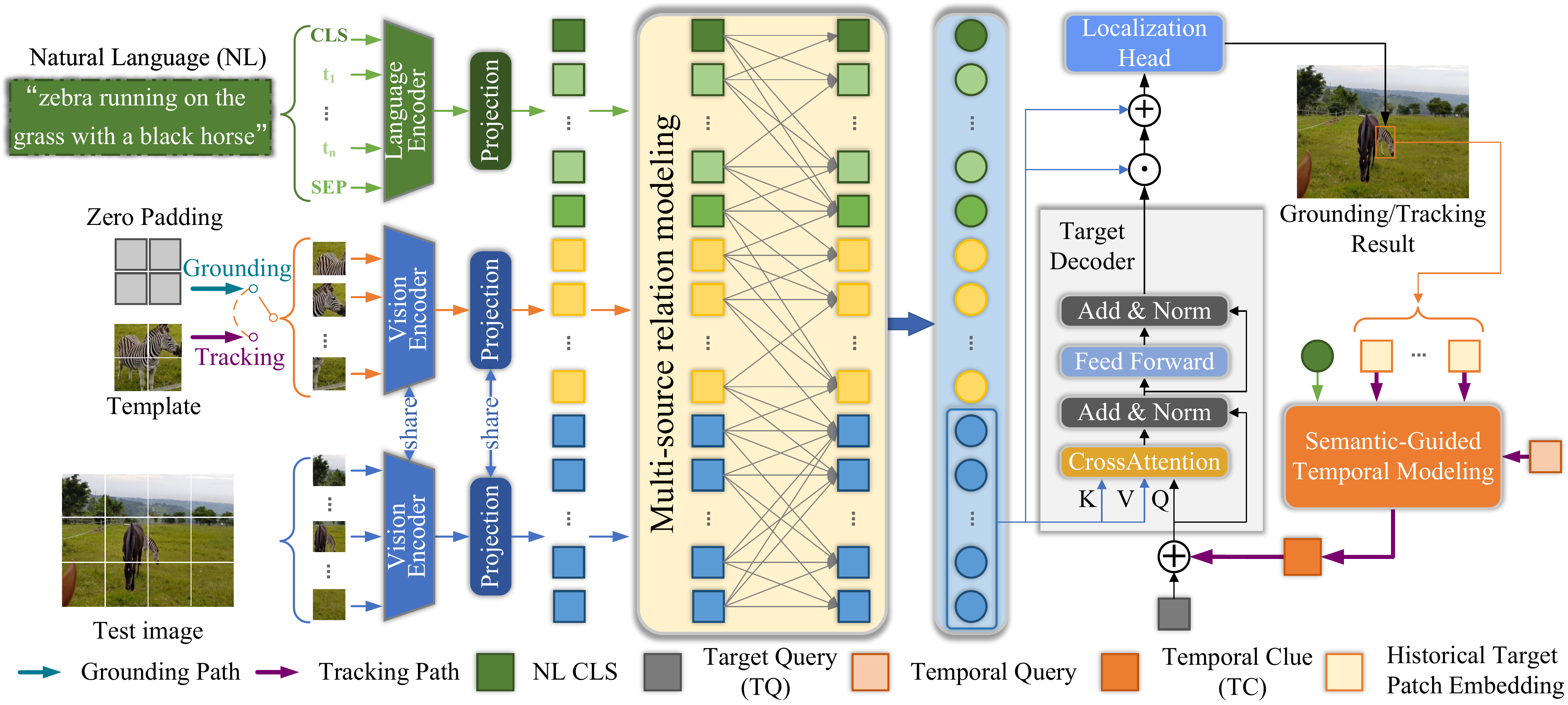}
\vspace{-2mm}
\caption{
Overview of our joint visual grounding and tracking framework. Given a sequence and a natural language description, we first feed the description, the first frame (test image), and zero padding tokens into the model for visual grounding and accordingly obtain the template image. For each subsequent frame (test image), we feed it with the description and template image into the model for tracking.
$\odot$ and $\oplus$ denote the element-wise product and summation operations, respectively.}
\label{Fig:framework}
\vspace{-2mm}
\end{figure*}

To this end, we propose a joint visual grounding and tracking framework for tracking by natural language specification, as shown in Figure~\ref{Fig:introduction}(b). Specifically, we look at these two tasks from a unified perspective and reformulate them as a unified one: localizing the referred target according to the given visual-language references. 
For visual grounding, the reference information is the natural language, while for visual tracking, the reference information is the natural language and historical target patch (usually called template). 
Thus, the crux of this unified task is to model the multi-source relations between the input references and the test image, which involve the cross-modality (visual and language) relation and the cross-time (historical target patch and current search image) relation. 
To deal with this issue, we introduce a transformer-based multi-source relation modeling module, which is flexible enough to accommodate the different references for grounding and tracking, to model the above relations effectively. It allows our method to switch between grounding and tracking according to different inputs. 

In addition, to improve the adaptability to the variations of the target, we resort to the historical prediction as they provide the temporal clue about the recent target appearance and propose a temporal modeling module to achieve this purpose. 
Considering that the natural language specification contains the global semantic information of the target, we use it as guidance to assist the temporal modeling module to focus on the target region instead of the noise in the previous prediction results. 

To conclude, we make the following contributions: (1) we propose a joint visual grounding and tracking framework for tracking by natural language specification, which unifies tracking and grounding as a unified task and can accommodate the different references of the grounding and tracking processes; (2) we propose a semantics-guided temporal modeling module to provide a temporal clue based on historical predictions for our joint model, which improves the adaptability of our method to the appearance variations of the target; (3) we achieve favorable performance against state-of-the-art algorithms on three natural language tracking datasets and one visual grounding dataset, which demonstrates the effectiveness of our approach.

\section{Related work}
\subsection{Tracking with bounding box specification}
The classical visual trackers~\cite{ECO, DaSiam, DiMP, PrDiMP} initialize their tracking procedure based on the given bounding box in the first frame. 
Inspired by the success of transformer~\cite{vaswani2017attention} in recognition~\cite{dosovitskiy2020image} and detection~\cite{DETR,Deformerble-DETR}, most of the recent trackers~\cite{TransMeetTracker, stark, ostrack, GTELT, ToMP, CSWinTT} are developed based on the transformer structure and achieve impressive performance on many tracking benchmarks.
To improve the discriminative of the features, TrDiMP~\cite{TransMeetTracker}, TrTr~\cite{zhao2021trtr}, and TransT~\cite{chen2021transformer} use the self-attention in the transformer encoder for feature enhancement and the cross-attention in the transformer decoder for information propagation between the template feature and the search feature.
Differently, STARK~\cite{stark} employs an encoder-decoder transformer architecture to effectively capture the global feature dependencies of both spatial and temporal information in video sequences, which achieves impressive performance.

The above-mentioned methods employ a two-stage approach, separating the feature extraction and feature interaction processes.
MixFormer~\cite{cui2022mixformer} and OSTrack~\cite{ostrack} construct a one-stream tracking pipeline and achieve state-of-the-art tracking performance.
Despite the promising performance on many tracking benchmarks, classical visual trackers are still hard to be straightly applied in the real world due to the limitation of the manually specified initialization method. 

\subsection{Tracking with natural language description}
Inspired by the development of the visual grounding task, Li \etal~\cite{li2017tracking} define the task of tracking by natural language specification. Yang \etal~\cite{GTI} decompose the problem into three sub-tasks, \ie~grounding, tracking, and integration, and process each sub-task separably by three modules. Differently, Feng \etal~\cite{feng2020real} solve this task following the tracking-by-detection formulation, which utilizes natural language to generate global proposals on each frame for tracking.
To provide a specialized platform for the task of tracking by natural language specification, Wang \etal~\cite{TNL2K} release a new benchmark for natural language-based tracking named TNL2K and propose two baselines initialized by natural language and natural language with bounding boxes, respectively.
Li \etal~\cite{li2022cross} employ a target-specific retrieval module to localize the target, which is used to initialize a local tracker.
Previous works either employ the off-the-shelf visual grounding model~\cite{2019onestagevg} or use a dedicated grounding module~\cite{feng2020real, TNL2K} to detect the target, inevitably separating the connection between visual grounding and tracking.
In this work, we propose a unified visual grounding and tracking framework for the task of tracking by natural language specification, which can perform visual grounding or tracking conditioned on different inputs.

\section{Method}
In this section, we present the proposed joint visual grounding and tracking framework, which implements the two core steps, \ie~grounding and tracking, of tracking by natural language specification with a single end-to-end model. 
To this end, our framework is designed to accommodate the relation modeling between the test image and different input references of grounding and tracking. 

\subsection{Joint grounding and tracking framework}
Figure~\ref{Fig:framework} illustrates the proposed joint visual grounding and tracking framework, which mainly consists of the language and vision encoders, the multi-source relation modeling module, the target decoder, the semantic-guided temporal modeling module, and the localization head.

Given the input references and the test image, the language and vision encoders first embed them into the specific feature spaces, yielding the token embeddings of the input words or image patches. 
Then we adopt two linear projection layers to project the language and vision token embeddings to the latent spaces with the same dimension. 
After that, the projected embeddings are fed into the multi-source relation modeling module, which models the relations between the multi-source references and the test image to enhance the target information in the embeddings of the test image. 
On top of the enhanced embeddings of the test image, the target decoder followed by the localization head is used to predict the bounding box of the referred target. 
For the visual grounding task, in which no template is available, we use zero padding tokens as placeholders to fill in the missing embeddings corresponding to the template in the input of the multi-source relation modeling module. 
During visual tracking, the semantics-guided temporal modeling module will generate a temporal clue for the target decoder, enabling our model to exploit historical target states.

The tracking process with natural language specification using our framework can be summarized as: (1) At the first frame, our model takes as input the natural language, the zero padding tokens, and the test image (the first frame) to perform visual grounding, and then the target template can be obtained based on the grounding result; (2) At every subsequent frame, our model takes as input the natural language, the template, and the test image (also called the search image) to perform language-assisted visual tracking.

\subsection{Vision and language feature extraction}
\noindent\textbf{Language encoder.}
Since being proposed, the transformer~\cite{vaswani2017attention} model has achieved great success in natural language processing. 
Thus, we opt for the classic language transformer model BERT~\cite{devlin2018bert} as the language encoder in our framework.
Given a natural language query $L_q$, we first tokenize the sentences and append a CLS token and a SEP token at the beginning and end of the tokenized language query, respectively, yielding a sequence of tokens $T = \{ \text{CLS}, \bm{t}_1,\bm{t}_2, \cdots, \bm{t}_N,\text{SEP}\}$, where $N$ is the max length of the language query.
Then we feed the above sequence into the language encoder and obtain the language token embeddings $\bm{F}_l \in  \mathbb{R}^{C_l \times (N+2)}$, where $C_l=768$ is the dimension of the output embedding.

\noindent\textbf{Vision encoder.}
Similar to the language encoder, we also opt for the transformer model to learn the vision feature representation. 
In particular, we adopt the vanilla Swin-Transformer~\cite{SWIN} as our vision encoder for its excellent performance of image feature learning. 
Herein we only remain the first three stages of Swin-Transformer, as the output feature resolution of the last stage is too small. 
For the grounding process, given a test image $I_t \in \mathbb{R}^{3 \times H_t \times W_t}$, we feed it into the vision encoder and then flatten the output features to obtain the token embeddings $\bm{F}_t \in \mathbb{R}^{C_v \times L_g}$.
Similarly, for the tracking process, given the template image $I_z \in \mathbb{R}^{3 \times H_z \times W_z}$ and test image $I_t \in \mathbb{R}^{3 \times H_t \times W_t}$, we feed them into the vision encoder and the flatten the output features to get their token embeddings $\bm{F}_z \in \mathbb{R}^{C_v \times L_z}$ and $\bm{F}_t \in \mathbb{R}^{C_v \times L_t}$, respectively. Herein $C_v = 512$.

\subsection{Multi-source relation modeling}\label{Sec:MSRM}
Unifying the visual grounding and tracking sub-tasks together requires the algorithm to model the relations between the test image and the different references. 
Thus, we propose the multi-source relation modeling module, which accommodates the different references for grounding and tracking, to achieve cross-modality and cross-time relation modeling for tracking by the natural language specification. 
Considering the ability of the self-attention operation to capture the global dependencies, we exploit a transformer encoder that mainly stacks the self-attention layer to perform relation modeling instead of using complex cross-modality fusion or cross-correlation methods.

For visual grounding where the template embeddings are not available, we use zero padding tensors $\bm{P}_{o}$ as placeholders to fill the missing template embeddings. 
We also set the mask value for zero padding tensors as 1 to mask them during the computation of self-attention in case they pollute the other useful information. 
Before performing relation modeling, we first use the language and vision projection layers to process the corresponding token embeddings to unify the dimension of the token embeddings from different modalities. 
The projected language embeddings, template image embeddings, and test image embedding are denoted by $\bm{P}_l= [\bm{p}_l^1, \bm{p}_l^2, \dotsc, \bm{p}_l^{N_l}]$, $\bm{P}_z= [\bm{p}_z^1, \bm{p}_z^2, \dotsc, \bm{p}_z^{N_z}]$, $\bm{P}_t= [\bm{p}_t^1, \bm{p}_t^2, \dotsc, \bm{p}_t^{N_t}]$, respectively.
Then, we concatenate the embeddings of the references ($\bm{P}_l$ and $\bm{P}_{o}$ for grounding, and $\bm{P}_l$ and $\bm{P}_{z}$ for tracking) and the embeddings of the test image $\bm{P}_t$, and feed them into the transformer encoder to model the multi-source relation for grounding or tracking, which can be formulated as:
\begin{align}
\setlength{\abovedisplayskip}{3pt}
\setlength{\belowdisplayskip}{3pt}
\label{Eq: enc}
    [\bm{h}_l, \bm{h}_{o}, \bm{h}_t] &= \Phi_{enc}([\bm{P}_l,\bm{P}_{o},\bm{P}_{t}]);\\
    [\bm{h}_l, \bm{h}_z, \bm{h}_t] &= \Phi_{enc}([\bm{P}_l,\bm{P}_{z},\bm{P}_{t}]).
\end{align}
Herein the output embeddings corresponding to the test image, \ie~$\bm{h}_t$, will be further used for target decoding and bounding box prediction. 
Besides, we add learnable position embeddings to $[\bm{P}_l,\bm{P}_{o},\bm{P}_{t}]$ and $[\bm{P}_l,\bm{P}_{z},\bm{P}_{t}]$ to retain the positional information.

\subsection{Target decoder and localization head}
\noindent\textbf{Target decoder.}
To further enhance the target information in $\bm{h}_t$, we follow STARK~\cite{stark} and use a target decoder(TDec) and a Target Query(TQ) to learn the discriminative target embeddings. 
The target decoder takes $\bm{h}_t$ and the target query as inputs. 
For visual grounding, the target query is an offline learned embedding that contains the potential target information. 
For visual tracking, we add the temporal clue outputted by the semantics-guided temporal modeling module to the offline learned embedding to obtain a target query with recent target appearance priors.

\noindent\textbf{Localization head.}
To achieve the unification of grounding and tracking, we apply a shared localization head for bounding box prediction. 
Particularly, we use the localization head proposed in~\cite{stark} due to its great performance and efficiency. 
As shown in Figure~\ref{Fig:framework}, we first compute the similarity between the output of the target decoder and $\bm{h}_t$ token by token, and then we use a residual connection to enhance the target-related regions. Finally, the localization head is constructed to predict the target bounding box.

\begin{figure}[t]
\centering
\includegraphics[width=1.0\columnwidth]{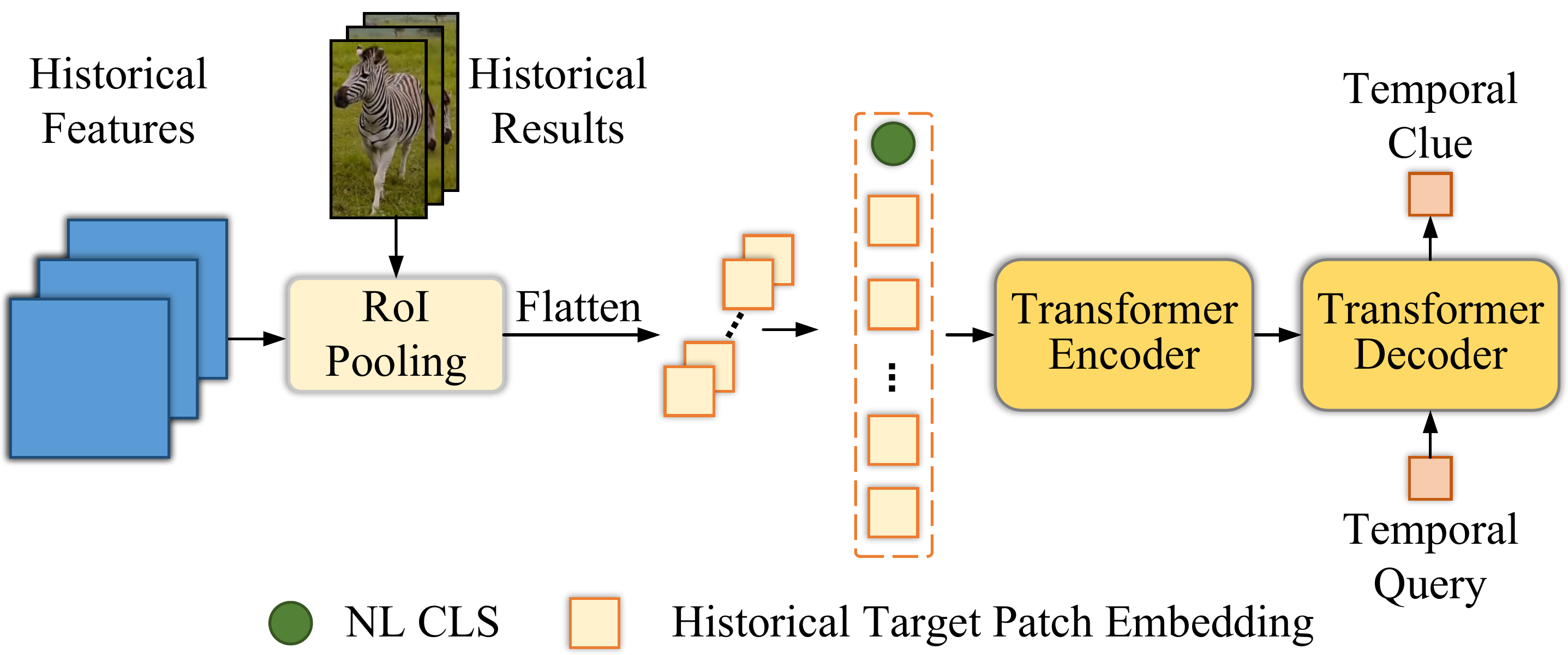}
\caption{Architecture of the proposed semantic-guided temporal modeling module.} 
\label{Fig:temporal_module}
\vspace{-2mm}
\end{figure}
\subsection{Semantics-guided temporal modeling}\label{Sec:SGTM}
Although grounding and tracking share the same relation modeling process, there still exist some differences between the two tasks. 
More concretely, grounding is performed on a standalone image but tracking is executed on consecutive video which means there exist historical appearances of the target for exploitation. 
To improve the adaptability to the variations of the target using the temporal information, we propose a Semantics-Guided Temporal Modeling (SGTM) module. It is designed to learn the temporal clue about recent target appearances from historical target states with semantics from the natural language as guidance.
Figure~\ref{Fig:temporal_module} shows the architecture of the SGTM module, which mainly consists of a transformer encoder and decoder. 

After the tracking process of each frame, we perform region of interest (RoI) pooling~\cite{fasterrcnn} on $\bm{h}_t$ according to the predicted box to obtain the target region features which are further flattened to obtain the historical target patch embeddings.
Besides, we also take into account the global semantic representation $\bm{h}_l^1$ of the Natural Language (NL) outputted by the multi-source relation modeling module, which is called the NL CLS token denoted as $\bm{h}_l^{CLS}$.
We directly concatenate the NL CLS token with the historical target patch embeddings and feed them into the encoder.
Herein, the encoder is used to enhance the token embeddings corresponding to the target region and suppress those corresponding to the noise region with the guidance of NL CLS.
After that, we utilize a decoder to compute cross-attention between enhanced target patch embeddings and a temporal query which is a learnable vector.
The temporal query can attend to the language context and historical target embeddings, thus learning a robust historical target representation for future tracking.

\subsection{End-to-end learning and inference}
\vspace{1mm}
\noindent\textbf{End-to-end learning.}
Every training sample is composed of a language query and a pair of images (a grounding patch and a tracking patch). During training, the forward propagation for one sample is divided into two steps: grounding and tracking.
In the grounding step, the network takes as input the language query and the grounding patch, and outputs the predicted box.
Then we crop the target region according to the predicted box of the grounding process as the template for the tracking step.
In the tracking step, the model takes as input the language query, the template, and the tracking patch (\ie~the search patch) and outputs the predicted boxes.
For each step, we use the GIoU~\cite{GIOU} Loss and L1 Loss to supervise the learning of the model, and the losses of the two steps are directly added.

\noindent\textbf{Inference.}
During inference, we perform grounding on the first frame, then accordingly crop the target region as the template.
Besides, we also generate the token embeddings of the template via RoI pooling and flatten operations, which are further fed into the SGTM module.
In each subsequent frame, we conduct tracking based on the language query, the template, and the temporal clue that SGTM learns.
Specifically, our model also can be initialized by the natural language and bounding box specifications together, which is validated to achieve better tracking performance.

\section{Experiments}
\newcommand{\tabincell}[2]{\begin{tabular}{@{}#1@{}}#2\end{tabular}}

\subsection{Experimental setup}
\noindent{\textbf{Implementation details.}}
We use Swin-B~\cite{SWIN} pre-trained on ImageNet~\cite{imagenet} as our vision encoder and use the base-uncased version of BERT~\cite{devlin2018bert} as our language encoder. 
For the vision input, the test image for grounding is resized such that its long edge is 320. No cropping or rotating is used to avoid breaking the alignment between the language and vision signal. The sizes of the template image and search image are set to $128\times128$ and $320\times320$, respectively. 
The size of historical target features for temporal modeling is set to $6\times 6$.
For the language input, the max length of the language is set to 40, including a CLS and a SEP token. 

We use the training splits of RefCOCOg-google~\cite{REFCOCO}, OTB99~\cite{li2017tracking}, LaSOT~\cite{LASOT}, and TNL2K~\cite{TNL2K} to train our model. 
We train our model for 300 epochs, and the warm-up strategy is adopted. 
The learning rates for the vision and language encoders and the other parameters first increase to $10^{-5}$ and $10^{-4}$ linearly at the first 30 epochs, respectively. 
The learning rates drop by 10 after the $200^{th}$ epoch every 50 epochs. 
We test the proposed algorithm on an NVIDIA 3090 GPU, and the tracking speed is about 39 FPS.

\noindent{\textbf{Datasets and metrics.}}
We evaluate our method for natural language tracking on the tracking benchmarks with natural language descriptions, including OTB99~\cite{li2017tracking}, LaSOT\cite{LASOT}, TNL2K~\cite{TNL2K}. 
All these datasets adopt success and precision to measure tracking performance. 
We report the Area Under the Curve (AUC) of the success plot and the precision score when the center location error threshold is set to 20 pixels.
Besides, we also evaluate our model for visual grounding on the Google-split val set of RefCOCOg~\cite{REFCOCO} using the standard protocol~\cite{2019onestagevg, transvg, VLTVG} and report the Top-1 accuracy. Herein the predicted box whose IoU with grounding truth is over 0.5 is considered as a correct prediction.

\begin{table}[t!]
\centering
\setlength{\tabcolsep}{8pt}
\renewcommand{\arraystretch}{1.05}
\caption{AUC and Precision (PRE) for four variants of our model on the LaSOT and TNL2K datasets.}
\vspace{-1mm}
\resizebox{0.8\linewidth}{!}{
\begin{tabular}{lccccc}
\toprule
\multirow{2}{*}{Variants} & \multicolumn{2}{c}{LaSOT} && \multicolumn{2}{c}{TNL2K} \\ 
\cline{2-3} \cline{5-6}  
                        & AUC         & PRE           && AUC         & PRE           \\ 
\midrule
SepRM                   & 0.518       & 0.512       && 0.491       & 0.471       \\
MSRM                    & 0.536       & 0.550       && 0.511       & 0.500       \\
MSRM-TDec               & 0.549       & 0.567       && 0.524       & 0.514       \\
MSRM-TM                 & 0.561       & 0.581       && 0.541       & 0.540       \\
\textbf{Our model}      & 0.569       & 0.593       && 0.546       & 0.550       \\ 
\bottomrule
\end{tabular}}
\label{Tab:ablation}
\end{table}

\begin{table}[t]
\centering
\setlength{\tabcolsep}{2.5pt}
\renewcommand{\arraystretch}{1.05}
\caption{Comparisons between separated and joint methods. We report FLOPs and inference time for grounding and tracking separately. Note that the time for all methods is tested on RTX 3090.}
\resizebox{1.0\linewidth}{!}{
\begin{tabular}{lcccccc}
\toprule
\multicolumn{2}{c}{\multirow{2}{*}{}} & \multicolumn{3}{c}{Separated Model} &\!& Joint Model \\
\multicolumn{2}{c}{}  & VLTVG+STARK    & VLTVG+OSTrack  & \multicolumn{1}{c}{SepRM} && Ours \\ 
\midrule
\multirow{2}{*}{FLOPs} & Grounding & 39.6G & 39.6G & 34.7G      && 34.9G \\
                       & Tracking & 20.4G & 48.3G & 38.5G       && 42.0G  \\ 
\midrule
\multirow{2}{*}{Time} & Grounding & 28.2ms & 28.2ms &  26.4ms     && 34.8ms \\
                      &Tracking   & 22.9ms & 8.3ms  &  20.6ms     && 25.3ms \\ 
\midrule
Params &  Total  & 169.8M & 214.7M &214.4M  && 153.0M \\ 
\midrule
\multicolumn{1}{l}{\multirow{2}{*}{AUC}} & LaSOT   & 0.446  & 0.524  &0.518  && 0.569  \\
\multicolumn{1}{l}{}                       & TNL2K  & 0.373  & 0.399  &0.491 && 0.546 \\
\bottomrule
\end{tabular}}
\label{Tab:flops_comparsion}
\end{table}

\subsection{Ablation study}
We first conduct experiments to analyze the effect of each main component in our model. To this end, we perform ablation studies on five variants of our model:

\noindent1) \textbf{SepRM}, which deploys \textbf{Sep}arated \textbf{R}elation \textbf{M}odeling modules, i.e., two independent transformer encoders, to perform visual grounding and tracking. In this variant, the target decoder is removed, and two localization heads are directly constructed on the outputs of the two separated relation modeling modules, respectively.

\noindent2) \textbf{MSRM}, which replaces the separated relation modeling modules with the \textbf{M}ulti-\textbf{S}ource \textbf{R}elation \textbf{M}odeling module in SepRM, and only deploys a single localization head.

\noindent3) \textbf{MSRM-TDec}, which introduces a \textbf{T}arget \textbf{Dec}oder into the MSRM variant.

\noindent4) \textbf{MSRM-TM}, which introduces our proposed semantic-guided temporal modeling into the MSRM-TDec variant but only the historical target patches as inputs.

\noindent5) \textbf{Our model}, which introduces the NL CLS tokens into SGTM module, further exploits language information to improve the adaptability to the target appearance variations.

Table~\ref{Tab:ablation} presents the experimental results of these variants on LaSOT~\cite{LASOT} and TNL2K~\cite{TNL2K}.

\noindent\textbf{Effect of multi-source relation modeling.}
The performance gap between SepRM and MSRM demonstrates the effectiveness of multi-sources relation modeling to embed the reference information from the natural language and template into the feature of the test image for both grounding and tracking.

\noindent\textbf{Effect of target encoder.}
The performance comparison between MSRM and MSRM-TDec shows that the target decoder plays an important role in decoding the target state information from the test feature after relation modeling. 
Similar performance gains from the target decoder are also observed in STARK~\cite{stark}. 

\noindent\textbf{Effect of temporal modeling.}
A large performance boost arises at the comparison from MSRM-TDec to our intact model, which is benefited from temporal modeling. 
The temporal modeling module provides effective temporal clues for the decoder to adapt to the appearance variations of the target, thus greatly improving tracking performance.

\noindent\textbf{Effect of semantics-guided temporal modeling.}
We can observe that incorporating the NL CLS token brings performance gains, especially in the precision metric.
With the NL CLS token, our model predicts the target location more precisely, demonstrating the effectiveness of the language information in SGTM.

We further compare our joint visual grounding and tracking model with the separated methods in FLOPs, inference time, params, and AUC, as shown in Table~\ref{Tab:flops_comparsion}. Since the codes of existing separated methods have not been released, we construct two separated methods using state-of-the-art grounding (VLTVG~\cite{VLTVG}) and tracking (STARK~\cite{stark} and OSTrack-384~\cite{ostrack}) models.
Our proposed variant SepRM is also involved in the comparison. Compared with the separated methods, our approach performs comparably in computation, speed, and params, while performing substantially better in AUC on LaSOT and TNL2K.

\begin{table}[t]
\centering
\caption{AUC and Precision (PRE) of different methods on the OTB99, LaSOT, and TNL2K datasets. The best and second-best results are marked in \textbf{bold} and \underline{underline}. BB and NL denote the Bounding Box and Natural Language, respectively.}
\vspace{-1mm}
\setlength{\tabcolsep}{3.5pt}
\renewcommand{\arraystretch}{1.05}
\label{Tab:Tracking_results}
\resizebox{1.0\linewidth}{!}{
\begin{tabular}{lcccccccc}
\toprule
\multirow{2}{*}{Algorithms}  & \multirow{2}{*}{Initialize} & OTB99       & LaSOT    & TNL2K \\  
& & AUC \textbar\ \ PRE  & AUC \textbar\ \  PRE & AUC \textbar \ \ PRE \\
\midrule
AutoMatch     & BB         & --         & 0.583 \textbar\ 0.599 & 0.472 \textbar\ 0.435 \\
TrDiMP       & BB    & -- & 0.639 \textbar\  0.663 & \underline{0.523} \textbar\ \textbf{0.528} \\
TransT        & BB    & -- & 0.649 \textbar\ 0.690 & 0.507 \textbar\ 0.517 \\
STARK      & BB    & -- & 0.671 \textbar\ 0.712 & --      \\
KeepTrack     & BB    & -- & 0.671 \textbar\ 0.702 & --        \\
SwinTrack-B & BB    & -- & \underline{0.696}  \textbar\  \underline{0.741} & --       \\
OSTrack-384   & BB    & -- & \textbf{0.711} \textbar\ \textbf{0.776} & \textbf{0.559} \textbar\ \  \ --\ \   \\ \midrule
TNLS-II     & NL      & 0.250 \textbar\ 0.290     & --     & --        \\
RTTNLD   & NL      & 0.540 \textbar\ \textbf{0.780}     & 0.280 \textbar\ 0.280 & --       \\
GTI          & NL      & \underline{0.581} \textbar\ 0.732     & 0.478 \textbar\ 0.476 & --   \\
TNL2K-1       & NL      & 0.190 \textbar\ 0.240     & 0.510 \textbar\ 0.490 & 0.110 \textbar\ 0.060  \\
CTRNLT   & NL      & 0.530 \textbar\ 0.720     & \underline{0.520} \textbar\  \underline{0.510} & \underline{0.140}  \textbar\  \underline{0.090}  \\
\textbf{Ours}              & NL      & \textbf{0.592}  \textbar\  \underline{0.776}  & \textbf{0.569}  \textbar\  \textbf{0.593} & \textbf{0.546}  \textbar\  \textbf{0.550} \\ \midrule
TNLS-III     & NL+BB & 0.550  \textbar\  0.720     & --    & --     \\
RTTNLD   & NL+BB & 0.610  \textbar\  0.790     & 0.350 \textbar\ 0.350 & 0.250  \textbar\  0.270 \\
TNL2K-2      & NL+BB & \underline{0.680} \textbar\ \underline{0.880}     & 0.510 \textbar\ 0.550 & 0.420 \textbar\ 0.420 \\
SNLT        & NL+BB & 0.666 \textbar\ 0.804     & 0.540 \textbar\ 0.576 & 0.276 \textbar\ 0.419 \\
VLTTT         & NL+BB & \textbf{0.764} \textbar\ \textbf{0.931}     & \textbf{0.673} \textbar\ \textbf{0.721} & \underline{0.531} \textbar\ \underline{0.533} \\
\textbf{Ours}             & NL+BB & 0.653 \textbar\  0.856          & \underline{0.604} \textbar\ \underline{0.636}      & \textbf{0.569} \textbar\ \textbf{0.581}    \\ \bottomrule
\end{tabular}}
\vspace{-2mm}
\end{table}
\subsection{Tracking by only language specification}
In this section, we compare our approach with state-of-the-art trackers using only language for initialization including TNLS-II~\cite{li2017tracking}, RTTNLD~\cite{feng2020real}, GTI~\cite{GTI}, TNL2K-1~\cite{TNL2K}, CTRNLT~\cite{li2022cross} on TNL2K~\cite{TNL2K}, LaSOT~\cite{LASOT}, and OTB99~\cite{li2017tracking}.  
Table~\ref{Tab:Tracking_results} reports the experimental results of different algorithms on these three datasets.
The results of the classical trackers (AutoMatch~\cite{AutoMatch}, TiDiMP~\cite{TransMeetTracker}, TransT~\cite{chen2021transformer}, STARK~\cite{stark}, KeepTrack~\cite{keeptrack}, SwinTrack-B~\cite{swintrack}, and OSTrack-384\cite{ostrack}) are also presented.

\vspace{1mm}
\noindent\textbf{TNL2K.}
Compared with the TNL2K-1 and CTRNLT methods that use separated grounding and tracking models, our method achieves substantial performance gains of 40.6\% and 46.0\% in AUC, precision respectively, which manifests the effectiveness of the proposed joint visual grounding and tracking framework. 
Besides, our method also performs marginally better than the VLTTT method that uses both natural language and a bounding box for initialization on the TNL2K dataset.

\noindent\textbf{LaSOT.}
Compared with CTRNLT, our method improves tracking performance by 4.9\% in AUC and 8.3\% in precision. Besides, our approach also achieves substantial performance gains in comparison with TNL2K-1. 

\noindent\textbf{OTB99.} 
Compared with recently proposed state-of-the-art natural language trackers, our method achieves better performance (an AUC score of 59.2\%) on OTB99, which validates the effectiveness of our method.

\begin{figure}[t]
\centering
\includegraphics[width=1.0\columnwidth]{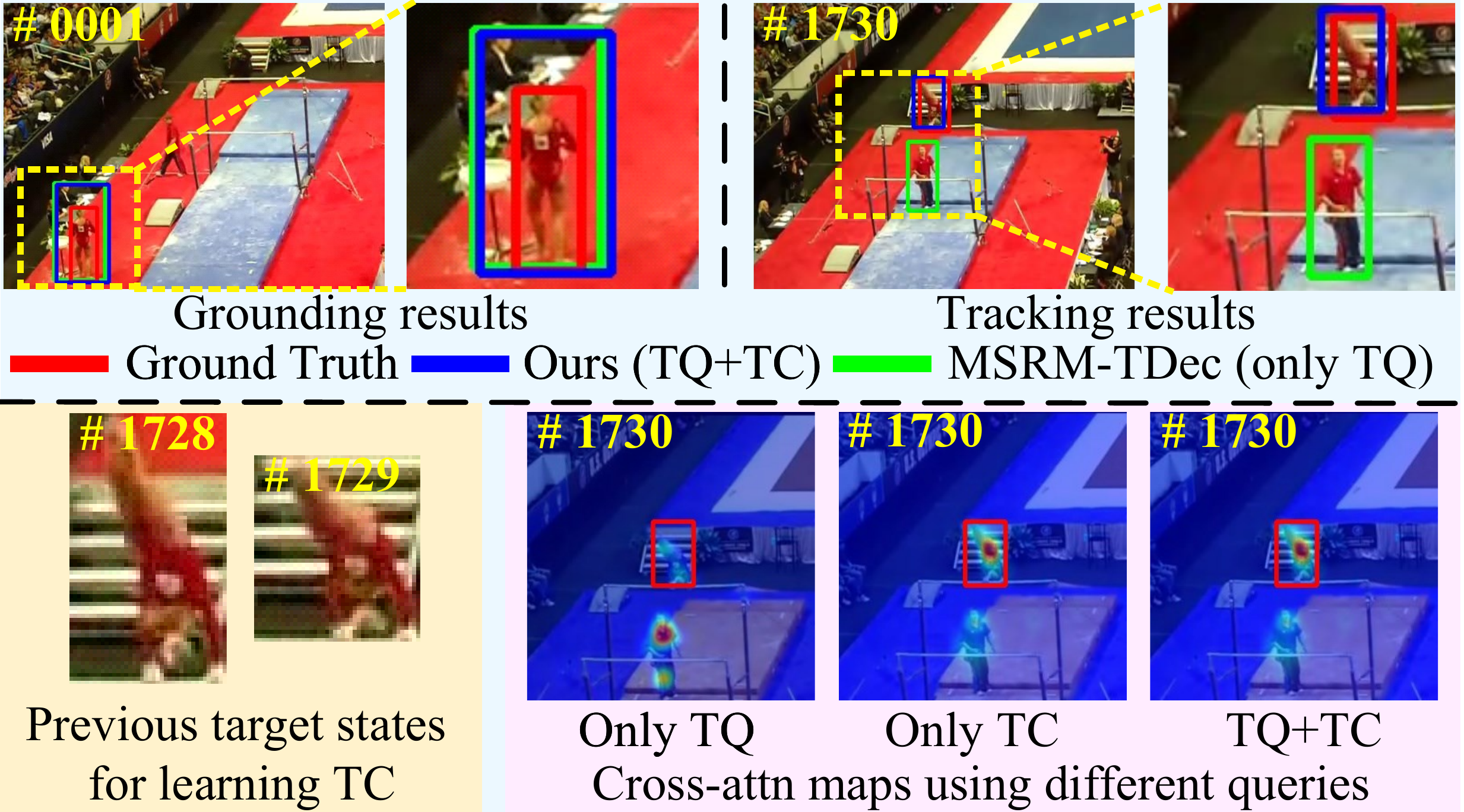}
\caption{Visualization for revealing what temporal clue SGTM learns. 
``Only TQ'' denotes only Target Query is used as the query in the target decoder. 
``Only TC'' denotes only Temporal Clue is used as the query. 
``TQ+TC'' denotes the summation of the Target Query and Temporal Clue is used as the query.
}
\label{Fig:vis_temporal}
\end{figure}
\noindent\textbf{Qualitative study.} 
To further investigate the effectiveness of the SGTM module, we visualize the cross-attention maps in the target decoder and the prediction results in Figure~\ref{Fig:vis_temporal}.
The attention map w/o TC (only TQ) has high responses on the distractor, and the variant MSRM-TDec loses the target, as the target appearance varies a lot compared to the template. 
By contrast, the learned TC attends to the real target, and it helps our model (TQ+TC) focus on the real target and predicts a precise bounding box.

\begin{figure*}[t]
\centering
\includegraphics[width=0.965\textwidth]{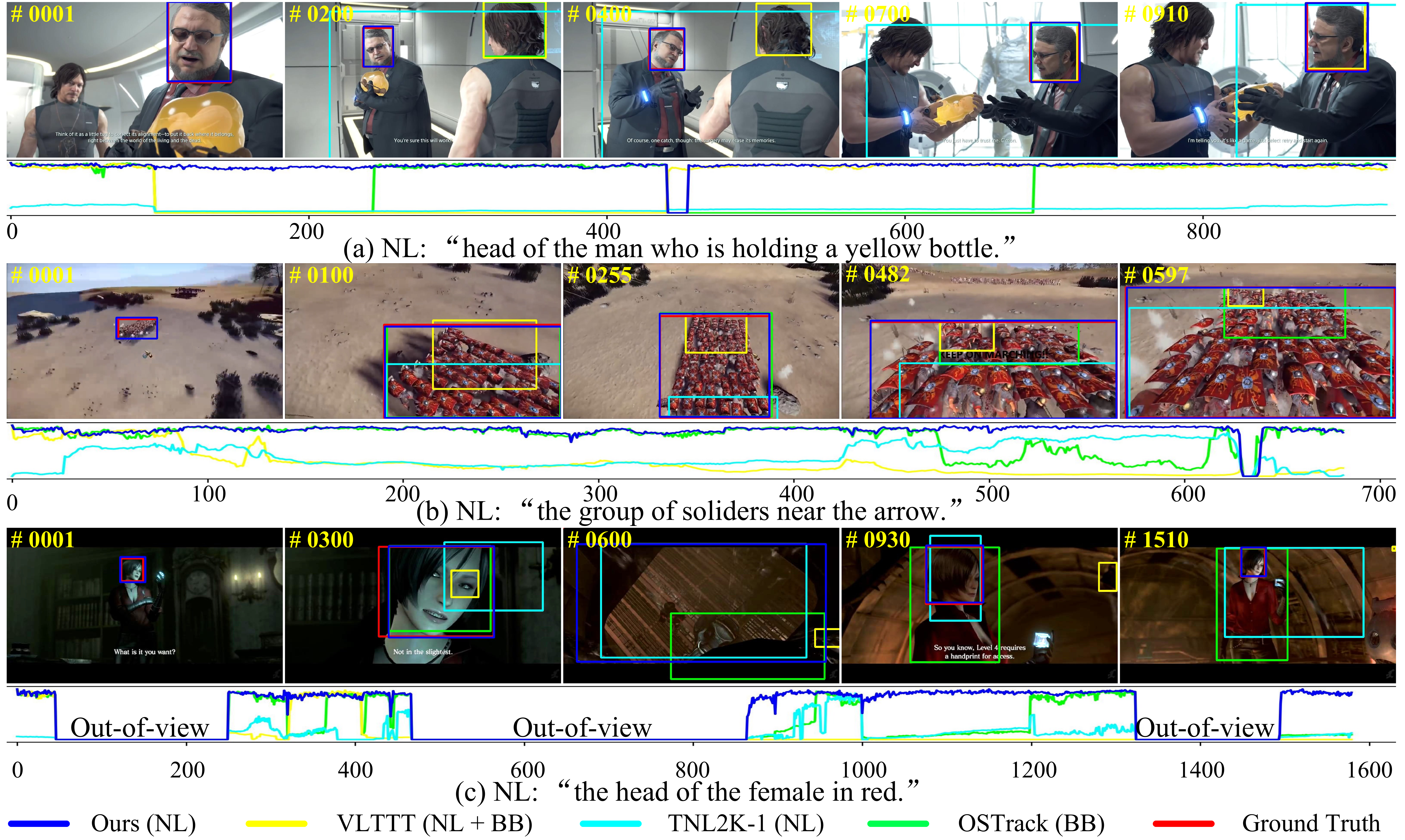}
\vspace{-1mm}
\caption{Qualitative comparison on three challenging sequences. From top to bottom, the main challenge factors are viewpoint change, appearance variation, and out-of-view, respectively. Our model is more robust than other trackers.}
\label{Fig:qualitative}
\end{figure*}

As shown in Figure~\ref{Fig:qualitative}, we visualize the tracking results on three challenging sequences in TNL2K, in which the main challenges are: viewpoint change, appearance variation, and out-of-view, respectively. 
We can observe that our model is more robust than other trackers. 
For example, in the second sequence, TNL2K-1 cannot predict the target box accurately at the beginning, and VLTTT and OSTrack gradually predict inaccurate target boxes as the target appearance varies. 
By contrast, our method predicts the target box more accurately.
These comparisons validate the effectiveness of the SGTM module.

\begin{figure*}[t]
\centering
\includegraphics[width=0.965\textwidth]{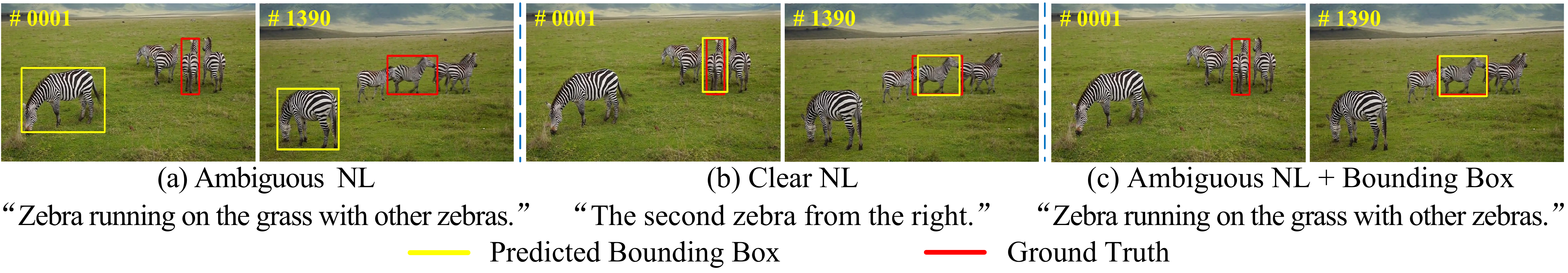}
\vspace{-1mm}
\caption{Analysis about the effect of ambiguous natural language (NL) on a zebra sequence. 
Given the original NL description (a) with ambiguity from LaSOT, our method localizes the wrong target at the first frame and consequently fails in the whole sequence. 
By contrast, given a clear NL description (b) or providing a bounding box (c) to eliminate ambiguity, our method can successfully locate the target.
} 
\label{Fig:casestudy2}
\vspace{-1mm}
\end{figure*}

\begin{table}[t]
\centering
\setlength{\tabcolsep}{3pt}
\renewcommand{\arraystretch}{1.15}
\caption{Comparison of our method with state-of-the-art algorithms for visual grounding on the val set of RefCOCOg.}
\vspace{-1mm}
\resizebox{1.0\linewidth}{!}{
\begin{tabular}{lcccccc}
\toprule
Algorithms & NMTree & LBYL-Net & ReSC-Large & TransVG & VLTVG  & Ours   \\
&~\cite{NMTree} & ~\cite{LBYL-Net}& ~\cite{ReSC-Large} &~\cite{transvg} & ~\cite{VLTVG} \\
\midrule
Accuracy & 0.6178 & 0.6270   & 0.6312     & 0.6702  & \textbf{0.7298} & \underline{0.7007} \\
\bottomrule
\end{tabular}}
\label{Tab:groundingresult}
\vspace{-1mm}
\end{table}

\subsection{Tracking by language and box specification}
We also compare our approach with state-of-the-art trackers in the same settings that use both natural language and a bounding box for initialization. 
The trackers involved in comparisons include TNLS-III~\cite{li2017tracking}, RTTNLD~\cite{feng2020real}, TNL2K-2~\cite{TNL2K}, SNLT~\cite{feng2021siamese}, VLTTT~\cite{guo2022divert}. Table~\ref{Tab:Tracking_results} presents the experimental results of these methods.

\noindent\textbf{TNL2K.}
Compare with VLTTT, our model improves performance by 3.8\% and 4.8 \% in AUC and precision, respectively, demonstrating the effectiveness of our method.

\noindent\textbf{LaSOT.}
The proposed method obtains an AUC score of 0.604 and a precision score of  0.636 on LaSOT.
The VLTTT method is designed based on the classical tracker TransT~\cite{chen2021transformer} that already achieves a favorable performance on LaSOT and performs better than our method. 
Compared to the other trackers with language and box specifications, our method achieves substantial performance gains.

\noindent\textbf{OTB99.}
VLTTT achieves the best performance with an AUC score of 0.653 and a precision score of 0.856, which is higher than ours. 
Nevertheless, our method performs on par with SNLT and TNL2K-2.

\subsection{Performance on visual grounding}\label{grounding}
We also report the visual grounding performance of our framework on the val set of RefCOCOg split by Google, as shown in Table~\ref{Tab:groundingresult}.
Although not originally designed for visual grounding, our method performs comparably with the state-of-the-art visual grounding algorithm VLTVG~\cite{VLTVG}.

\subsection{Limitations}
Herein we discuss the limitations of our algorithm. 
Our method is designed for tracking the target of interest only based on the natural language description. 
Inevitably, it is sensitive to ambiguous natural language descriptions to some extent. As shown in Figure~\ref{Fig:casestudy2}(a), given the ambiguous description ``Zebra running on the grass with other zebras" that does not clearly specify which zebra is the target one, our method fails to locate the target zebra at the beginning. 
One solution for this issue is to provide a clear description. 
As shown in Figure~\ref{Fig:casestudy2}(b), given a clear location description of the target zebra, our algorithm successfully locates the target and keeps tracking it. 
Another alternative solution is to provide a bounding box for disambiguation, as shown in Figure~\ref{Fig:casestudy2}(c). With the correction of the additional bounding box, our method can also successfully track the target zebra.

\section{Conclusion}
We have presented a joint visual grounding and tracking framework to connect two tasks by unifying the relation modeling among the multi-source references and test image, which includes the cross-modality (visual and language) relation and cross-temporal (historical target patch and current search frame) relation.
Besides, we propose a semantics-guided temporal modeling module modeling the historical target states with global semantic information as guidance, which effectively improves tracking performance.
Our method achieves favorable performance against state-of-the-art algorithms on three natural language tracking datasets and one visual grounding dataset.

\vspace{1mm}
\noindent\textbf{Acknowledgement.}
This study was supported in part by the National Natural Science Foundation of China (No. 62172126), the Shenzhen Research Council (No. JCYJ20210324120202006), and the Key Research Project of Peng Cheng Laboratory (No. PCL2021A07).

{\small
\bibliographystyle{ieee_fullname}
\bibliography{egbib}

\begin{thebibliography}{10}\itemsep=-1pt

\bibitem{DiMP}
Goutam Bhat, Martin Danelljan, Luc~Van Gool, and Radu Timofte.
\newblock Learning discriminative model prediction for tracking.
\newblock In {\em ICCV}, pages 6182--6191, 2019.

\bibitem{DETR}
Nicolas Carion, Francisco Massa, Gabriel Synnaeve, Nicolas Usunier, Alexander
  Kirillov, and Sergey Zagoruyko.
\newblock End-to-end object detection with transformers.
\newblock In {\em ECCV}, pages 213--229, 2020.

\bibitem{chen2021transformer}
Xin Chen, Bin Yan, Jiawen Zhu, Dong Wang, Xiaoyun Yang, and Huchuan Lu.
\newblock Transformer tracking.
\newblock In {\em CVPR}, pages 8126--8135, 2021.

\bibitem{cui2022mixformer}
Yutao Cui, Cheng Jiang, Limin Wang, and Gangshan Wu.
\newblock Mixformer: End-to-end tracking with iterative mixed attention.
\newblock In {\em CVPR}, pages 13608--13618, 2022.

\bibitem{ECO}
Martin Danelljan, Goutam Bhat, Fahad Shahbaz~Khan, and Michael Felsberg.
\newblock Eco: Efficient convolution operators for tracking.
\newblock In {\em CVPR}, pages 6638--6646, 2017.

\bibitem{PrDiMP}
Martin Danelljan, Luc~Van Gool, and Radu Timofte.
\newblock Probabilistic regression for visual tracking.
\newblock In {\em CVPR}, pages 7183--7192, 2020.

\bibitem{transvg}
Jiajun Deng, Zhengyuan Yang, Tianlang Chen, Wengang Zhou, and Houqiang Li.
\newblock Transvg: End-to-end visual grounding with transformers.
\newblock In {\em ICCV}, pages 1769--1779, 2021.

\bibitem{devlin2018bert}
Jacob Devlin, Ming-Wei Chang, Kenton Lee, and Kristina Toutanova.
\newblock Bert: Pre-training of deep bidirectional transformers for language
  understanding.
\newblock {\em arXiv preprint arXiv:1810.04805}, 2018.

\bibitem{dosovitskiy2020image}
Alexey Dosovitskiy, Lucas Beyer, Alexander Kolesnikov, Dirk Weissenborn,
  Xiaohua Zhai, Thomas Unterthiner, Mostafa Dehghani, Matthias Minderer, Georg
  Heigold, Sylvain Gelly, et~al.
\newblock An image is worth 16x16 words: Transformers for image recognition at
  scale.
\newblock {\em arXiv preprint arXiv:2010.11929}, 2020.

\bibitem{LASOT}
Heng Fan, Liting Lin, Fan Yang, Peng Chu, Ge Deng, Sijia Yu, Hexin Bai, Yong
  Xu, Chunyuan Liao, and Haibin Ling.
\newblock Lasot: A high-quality benchmark for large-scale single object
  tracking.
\newblock In {\em CVPR}, pages 5374--5383, 2019.

\bibitem{feng2020real}
Qi Feng, Vitaly Ablavsky, Qinxun Bai, Guorong Li, and Stan Sclaroff.
\newblock Real-time visual object tracking with natural language description.
\newblock In {\em WACV}, pages 700--709, 2020.

\bibitem{feng2021siamese}
Qi Feng, Vitaly Ablavsky, Qinxun Bai, and Stan Sclaroff.
\newblock Siamese natural language tracker: Tracking by natural language
  descriptions with siamese trackers.
\newblock In {\em CVPR}, pages 5851--5860, 2021.

\bibitem{guo2022divert}
Mingzhe Guo, Zhipeng Zhang, Heng Fan, and Liping Jing.
\newblock Divert more attention to vision-language tracking.
\newblock {\em arXiv preprint arXiv:2207.01076}, 2022.

\bibitem{LBYL-Net}
Binbin Huang, Dongze Lian, Weixin Luo, and Shenghua Gao.
\newblock Look before you leap: Learning landmark features for one-stage visual
  grounding.
\newblock In {\em CVPR}, pages 16888--16897, 2021.

\bibitem{imagenet}
Alex Krizhevsky, Ilya Sutskever, and Geoffrey~E Hinton.
\newblock Imagenet classification with deep convolutional neural networks.
\newblock {\em Communications of the ACM}, 60(6):84--90, 2017.

\bibitem{li2019siamrpn++}
Bo Li, Wei Wu, Qiang Wang, Fangyi Zhang, Junliang Xing, and Junjie Yan.
\newblock Siamrpn++: Evolution of siamese visual tracking with very deep
  networks.
\newblock In {\em CVPR}, pages 4282--4291, 2019.

\bibitem{li2022cross}
Yihao Li, Jun Yu, Zhongpeng Cai, and Yuwen Pan.
\newblock Cross-modal target retrieval for tracking by natural language.
\newblock In {\em CVPR}, pages 4931--4940, 2022.

\bibitem{li2017tracking}
Zhenyang Li, Ran Tao, Efstratios Gavves, Cees~GM Snoek, and Arnold~WM
  Smeulders.
\newblock Tracking by natural language specification.
\newblock In {\em CVPR}, pages 6495--6503, 2017.

\bibitem{swintrack}
Liting Lin, Heng Fan, Yong Xu, and Haibin Ling.
\newblock Swintrack: A simple and strong baseline for transformer tracking.
\newblock {\em arXiv preprint arXiv:2112.00995}, 2021.

\bibitem{NMTree}
Daqing Liu, Hanwang Zhang, Feng Wu, and Zheng-Jun Zha.
\newblock Learning to assemble neural module tree networks for visual
  grounding.
\newblock In {\em ICCV}, pages 4673--4682, 2019.

\bibitem{SWIN}
Ze Liu, Yutong Lin, Yue Cao, Han Hu, Yixuan Wei, Zheng Zhang, Stephen Lin, and
  Baining Guo.
\newblock Swin transformer: Hierarchical vision transformer using shifted
  windows.
\newblock In {\em ICCV}, pages 10012--10022, 2021.

\bibitem{REFCOCO}
Junhua Mao, Jonathan Huang, Alexander Toshev, Oana Camburu, Alan~L Yuille, and
  Kevin Murphy.
\newblock Generation and comprehension of unambiguous object descriptions.
\newblock In {\em CVPR}, pages 11--20, 2016.

\bibitem{ToMP}
Christoph Mayer, Martin Danelljan, Goutam Bhat, Matthieu Paul, Danda~Pani
  Paudel, Fisher Yu, and Luc Van~Gool.
\newblock Transforming model prediction for tracking.
\newblock In {\em CVPR}, pages 8731--8740, 2022.

\bibitem{keeptrack}
Christoph Mayer, Martin Danelljan, Danda~Pani Paudel, and Luc Van~Gool.
\newblock Learning target candidate association to keep track of what not to
  track.
\newblock In {\em ICCV}, pages 13444--13454, 2021.

\bibitem{TrackingNet}
Matthias Muller, Adel Bibi, Silvio Giancola, Salman Alsubaihi, and Bernard
  Ghanem.
\newblock Trackingnet: A large-scale dataset and benchmark for object tracking
  in the wild.
\newblock In {\em ECCV}, pages 300--317, 2018.

\bibitem{fasterrcnn}
Shaoqing Ren, Kaiming He, Ross Girshick, and Jian Sun.
\newblock Faster r-cnn: Towards real-time object detection with region proposal
  networks.
\newblock {\em NeurIPS}, 28, 2015.

\bibitem{GIOU}
Hamid Rezatofighi, Nathan Tsoi, JunYoung Gwak, Amir Sadeghian, Ian Reid, and
  Silvio Savarese.
\newblock Generalized intersection over union: A metric and a loss for bounding
  box regression.
\newblock In {\em CVPR}, pages 658--666, 2019.

\bibitem{CSWinTT}
Zikai Song, Junqing Yu, Yi-Ping~Phoebe Chen, and Wei Yang.
\newblock Transformer tracking with cyclic shifting window attention.
\newblock In {\em CVPR}, pages 8791--8800, 2022.

\bibitem{vaswani2017attention}
Ashish Vaswani, Noam Shazeer, Niki Parmar, Jakob Uszkoreit, Llion Jones,
  Aidan~N Gomez, {\L}ukasz Kaiser, and Illia Polosukhin.
\newblock Attention is all you need.
\newblock {\em NeurIPS}, 30, 2017.

\bibitem{TransMeetTracker}
Ning Wang, Wengang Zhou, Jie Wang, and Houqiang Li.
\newblock Transformer meets tracker: Exploiting temporal context for robust
  visual tracking.
\newblock In {\em CVPR}, pages 1571--1580, 2021.

\bibitem{wang2018describe}
Xiao Wang, Chenglong Li, Rui Yang, Tianzhu Zhang, Jin Tang, and Bin Luo.
\newblock Describe and attend to track: Learning natural language guided
  structural representation and visual attention for object tracking.
\newblock {\em arXiv preprint arXiv:1811.10014}, 2018.

\bibitem{TNL2K}
Xiao Wang, Xiujun Shu, Zhipeng Zhang, Bo Jiang, Yaowei Wang, Yonghong Tian, and
  Feng Wu.
\newblock Towards more flexible and accurate object tracking with natural
  language: Algorithms and benchmark.
\newblock In {\em CVPR}, pages 13763--13773, 2021.

\bibitem{OTB2013}
Yi Wu, Jongwoo Lim, and Ming-Hsuan Yang.
\newblock Online object tracking: A benchmark.
\newblock In {\em CVPR}, pages 2411--2418, 2013.

\bibitem{OTB2015}
Yi Wu, Jongwoo Lim, and Ming-Hsuan Yang.
\newblock Object tracking benchmark.
\newblock {\em IEEE TPAMI}, 37(9):1834--1848, 2015.

\bibitem{stark}
Bin Yan, Houwen Peng, Jianlong Fu, Dong Wang, and Huchuan Lu.
\newblock Learning spatio-temporal transformer for visual tracking.
\newblock In {\em ICCV}, pages 10448--10457, 2021.

\bibitem{VLTVG}
Li Yang, Yan Xu, Chunfeng Yuan, Wei Liu, Bing Li, and Weiming Hu.
\newblock Improving visual grounding with visual-linguistic verification and
  iterative reasoning.
\newblock In {\em CVPR}, pages 9499--9508, 2022.

\bibitem{ReSC-Large}
Zhengyuan Yang, Tianlang Chen, Liwei Wang, and Jiebo Luo.
\newblock Improving one-stage visual grounding by recursive sub-query
  construction.
\newblock In {\em ECCV}, pages 387--404, 2020.

\bibitem{2019onestagevg}
Zhengyuan Yang, Boqing Gong, Liwei Wang, Wenbing Huang, Dong Yu, and Jiebo Luo.
\newblock A fast and accurate one-stage approach to visual grounding.
\newblock In {\em ICCV}, pages 4683--4693, 2019.

\bibitem{GTI}
Zhengyuan Yang, Tushar Kumar, Tianlang Chen, Jingsong Su, and Jiebo Luo.
\newblock Grounding-tracking-integration.
\newblock {\em IEEE TCSVT}, 31(9):3433--3443, 2020.

\bibitem{ostrack}
Botao Ye, Hong Chang, Bingpeng Ma, and Shiguang Shan.
\newblock Joint feature learning and relation modeling for tracking: A
  one-stream framework.
\newblock In {\em ECCV}, pages 341--357, 2022.

\bibitem{yilmaz2006object}
Alper Yilmaz, Omar Javed, and Mubarak Shah.
\newblock Object tracking: A survey.
\newblock {\em Acm computing surveys}, 38(4):13--es, 2006.

\bibitem{AutoMatch}
Zhipeng Zhang, Yihao Liu, Xiao Wang, Bing Li, and Weiming Hu.
\newblock Learn to match: Automatic matching network design for visual
  tracking.
\newblock In {\em ICCV}, pages 13339--13348, 2021.

\bibitem{zhao2021trtr}
Moju Zhao, Kei Okada, and Masayuki Inaba.
\newblock Trtr: Visual tracking with transformer.
\newblock {\em arXiv preprint arXiv:2105.03817}, 2021.

\bibitem{GTELT}
Zikun Zhou, Jianqiu Chen, Wenjie Pei, Kaige Mao, Hongpeng Wang, and Zhenyu He.
\newblock Global tracking via ensemble of local trackers.
\newblock In {\em CVPR}, pages 8761--8770, 2022.

\bibitem{Deformerble-DETR}
Xizhou Zhu, Weijie Su, Lewei Lu, Bin Li, Xiaogang Wang, and Jifeng Dai.
\newblock Deformable detr: Deformable transformers for end-to-end object
  detection.
\newblock In {\em ICLR}, 2020.

\bibitem{DaSiam}
Zheng Zhu, Qiang Wang, Bo Li, Wei Wu, Junjie Yan, and Weiming Hu.
\newblock Distractor-aware siamese networks for visual object tracking.
\newblock In {\em ECCV}, pages 101--117, 2018.

\end{thebibliography}
}

\end{document}